\documentclass[journal]{IEEEtran}
\usepackage{amsmath,graphicx,amssymb,amsfonts,hyperref,algorithm,algpseudocode,booktabs}

\begin{document}
%
\title{LLM Online Spatial-temporal Signal Reconstruction  Under Noise}
%
%
%

\author{Yi Yan, \IEEEmembership{Student Member, IEEE,}
        Dayu Qin, \IEEEmembership{Student Member, IEEE,}
        and Ercan E. Kuruoglu, \IEEEmembership{Senior Member, IEEE}
\thanks{Yi Yan and Dayu Qin contributed equally. Yi Yan was affiliated with the Tsinghua-Berkeley Shenzhen Institute, Shenzhen International Graduate School, Tsinghua University, during the completion of this work. Dayu Qin and Ercan E. Kuruoglu are currently affiliated with the Tsinghua-Berkeley Shenzhen Institute, Shenzhen International Graduate School, Tsinghua University.
Corresponding author: Ercan E. Kuruoglu; e-mail: kuruoglu@sz.tsinghua.edu.cn.}
}

%
%

\markboth{Journal of \LaTeX\ Class Files,~Vol.~14, No.~8, August~2015}%
{Shell \MakeLowercase{\textit{et al.}}: Bare Demo of IEEEtran.cls for IEEE Journals}
%



\maketitle

\begin{abstract}
This work introduces the LLM Online Spatial-temporal  Reconstruction (LLM-OSR) framework, which integrates Graph Signal Processing (GSP) and Large Language Models (LLMs) for online spatial-temporal signal reconstruction. The LLM-OSR utilizes a GSP-based spatial-temporal signal handler to enhance graph signals and employs LLMs to predict missing values based on spatiotemporal patterns. The performance of LLM-OSR is evaluated on traffic and meteorological datasets under varying Gaussian noise levels. Experimental results demonstrate that utilizing GPT-4-o mini within the LLM-OSR is accurate and robust under Gaussian noise conditions. The limitations are discussed along with future research insights, emphasizing the potential of combining GSP techniques with LLMs for solving spatiotemporal prediction tasks.
\end{abstract}


%
\IEEEpeerreviewmaketitle

\section{Introduction}

\IEEEPARstart{R}{ecent} advancements in artificial intelligence have led to breakthroughs in many fields such as healthcare diagnostics \cite{acosta2022multimodal} and investment portfolio construction \cite{sonkavde2023forecasting}, culminating in the development of Large Language Models.
Large Language Models (LLMs) are a type of artificial intelligence model that is designed for natural language processing (NLP) tasks for its ability to understand and generate large-scale texts \cite{radford2019better}. 
BERT is a groundbreaking predecessor to modern LLMs by demonstrating the power of bidirectional transformers for natural language understanding \cite{devlin2018bert}.
Modern LLMs, such as GPT-3 \cite{floridi2020gpt}, GPT-4 \cite{waisberg2023gpt}, ERNIE \cite{sun2021ernie} and Kimi AI \cite{chen2024application}, are trained on datasets with billions of words and typically transformers architecture, to manipulate natural language \cite{sohail2023decoding}. 
Several applications of LLMs have been explored: LLMs enhance interactive machine translation by delivering high-quality initial translations, adapting efficiently to user feedback, and minimizing training costs \cite{navarro2023exploring}; LLMs enhance sentiment analysis by generating domain-specific weak labels and enabling efficient model distillation for practical applications. \cite{deng2023llms}. 
However, LLMs still exhibit certain limitations. 
For example, if the information provided to LLMs is insufficient or the prompt is misleading, inaccurate, or incomplete healthcare decisions can be made by LLMs which can lead to physical or psychological harm \cite{tang2023evaluating}.
Despite their strength in processing text-based information, LLMs remain limited in handling multivariate data structures, necessitating exploration into methods like Graph Signal Processing (GSP) and Graph Neural Networks (GNNs) to analyze and model such complex datasets effectively.

Graph-based methods provide a powerful framework for modeling and analyzing correlations in complex multi-variate data and have been applied in many fields, such as neurological disorders screening \cite{miraglia2022brain} and financial crisis prediction \cite{2024_Market_Qin}. 
There are some application scenarios that are better suited for graph methods compared with computer vision (CV) and natural language processing (NLP) approaches, for instance, social network analysis \cite{rostami2023community, yan_2023_BiSCNN}, traffic prediction \cite{FAN_2024_RGDAN}, and quantum computing \cite{xu2024quantum}. 
By exploiting the graph topology, interactions among multivariate data are captured along with the multi-variate data, offering task performance that outperforms non-graph algorithms. 
In addition, GSP with spectral approaches enables efficient representation and extraction of spectral patterns in addition to the spatial patterns seen in graphs and has a wide range of applications, including flaw detection in the wire-based directed energy deposition \cite{bevans2023monitoring} and electroencephalography signal processing \cite{sharma2024emerging}.
Graph-based methods are conventionally applied to static machine learning tasks such as classification, regression (on time-invariant data), and clustering, which involve data analysis without considering temporal changes \cite{Dong_Graph_ML_2020, kipf2016semi}. 
However, some particular tasks such as traffic prediction, climate modeling, and financial forecasting, rely heavily on capturing spatial-temporal dependencies. 
Spatial-temporal graph algorithms, GSP and GNNs, were designed to solve these kinds of time-varying tasks and have succeeded in dealing with such challenges by effectively modeling relationships and temporal dynamics in graph-structured data \cite{yu2017spatio,wang2022robust, yan_2022_sign, FAN_2024_RGDAN}.

Some recent studies have further explored the integration of GSP with Large Language Models (LLMs), revealing substantial advantages. Firstly, this integration potentially expands the application of LLMs by enabling them to process and analyze time-varying graph structure data, which allows LLMs to engage in graph-based reasoning across varied scenarios such as academic and e-commerce networks \cite{jin2024large}. Secondly, GSP can leverage LLMs to process time-varying, multivariate data from a text-based perspective, providing a novel angle for analyzing dynamic complex networks \cite{jin2024large}. Notably, LLMs on graphs perform well not by relying on data leakage but because they interpret graphs as languages, where the node label (signal) is deemed more crucial than the structure itself \cite{huang2023can_LLM_graph}.
Additionally, the InstructGLM framework demonstrates how LLMs can effectively represent graph structures through natural language for node classification in citation networks. T   his approach eliminates the need for complex GNN pipelines and unifies graph learning with natural language processing, showcasing the potential of LLMs in graph tasks \cite{ye2023natural}.


In our work, we introduced a novel method, the Large Language Model for Online Spatial-temporal Reconstruction (LLM-OSR) algorithm, which combines the strengths of the time-varying GSP method and LLMs for efficient and accurate reconstruction of missing signals in dynamic spatiotemporal complex networks. This integrated approach uses GSP to denoise or enhance signal features, ensuring high-quality input for LLM-based prediction tasks.
The combination of GSP with LLMs presents noteworthy advantages. 
Firstly, this integration has the potential to expand the application of LLMs by enabling them to process and analyze time-varying graph structure data. 
Secondly, GSP can leverage LLMs to process time-varying, multivariate data from a text-based perspective, providing a novel perspective for analyzing dynamic complex networks. 

There are 2 main contributions to our paper: 
\begin{itemize}
    \item We introduced the LLM-OSR algorithm, a novel approach to reconstructing spatial-temporal signals in an online manner by seamlessly combining GSP-based techniques with LLM-driven predictors. This innovative approach effectively reconstructs time-varying graph signals in the presence of noise and missing values within spatiotemporal data.
    
    \item The LLM-OSR employs a sophisticated reverse embedding approach to transform spatial-temporal signals on graphs into coherent and contextually meaningful natural language expressions, making the information readily interpretable and actionable by LLMs. 
\end{itemize}



%


Here is the organization of this paper. The preliminary knowledge is presented in Section~\ref{sec_preliminaries}.
Section~\ref{sec_methods} provides a detailed discussion of the LLM-OSR.
The experimental results and corresponding discussions are covered in Section~\ref{sec_experiments}.
The limitations and some potential future extensions of the proposed LLM-OSR are discussed in 
Section~\ref{sec_limitation} provides an in-depth discussion of the limitations of LLM-OSR and outlines potential directions for future research.
Finally, Section~\ref{sec_conclusion} concludes the paper.

\begin{figure*}[htb]
    \centering
    \includegraphics[trim={10 380 10 75},clip,width=\linewidth]{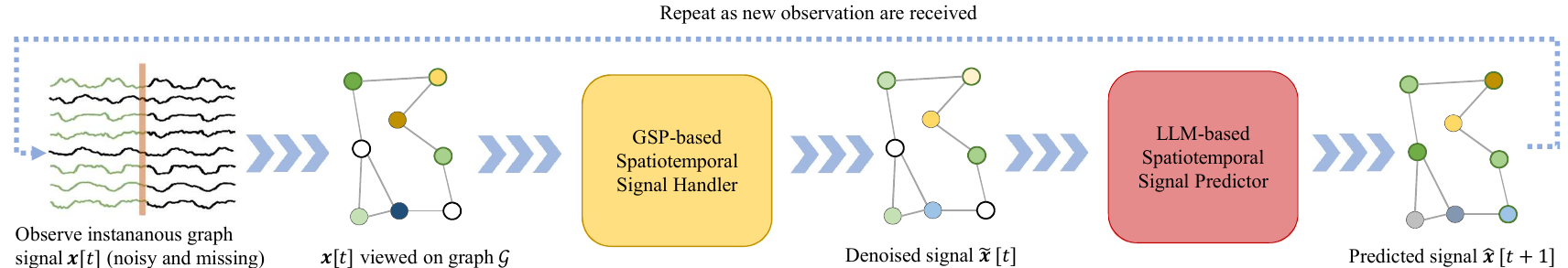}
    \caption{An overview of the LLM-OSR workflow}
    \label{fig_flow_overview}
\end{figure*}

\section{Preliminari Knowledge}
\label{sec_preliminaries}
\subsection{GSP Preliminaries}
In this paper, we consider an undirected and unweighted graph $\mathcal{G} = (\mathcal{V}, \mathcal{E})$, where $\mathcal{V} = \{v_1, \dots, v_i\}$ is the set of nodes or vertices, and $\mathcal{E}$ is the set of edges. 
We can represent the topology of the graph using the adjacency matrix $\mathbf{A}$, where $\mathbf{A} \in \mathbb{R}^{N \times N}$ and its elements are defined as follows:
\begin{equation}
    A_{ij} =
\begin{cases}
1, & \text{if there is an edge between } v_i \text{ and } v_j, \\
0, & \text{otherwise}.
\end{cases}
\label{eq_adjacency}
\end{equation}
The time-varying graph signal $\boldsymbol{x}[t]  \in \mathbb{R}^{N}$ is the multi-variate numerical data recorded on the graph nodes that change over time. 

In GSP, the spectral operations are defined using the graph Laplacian matrix $\mathbf{L} \in \mathbb{R}^{N \times N}$:
\begin{equation}
    \mathbf{L} = \mathbf{D} - \mathbf{A},
    \label{eq_laplacian}
\end{equation}
where $\mathbf{D}$ is the degree matrix, defined as $\mathbf{D} = \operatorname{diag}(\mathbf{1}^T \mathbf{A})$ and $\mathbf{1}$ is an all ones vector.
The spectral operations are conducted through GSP by having the GFT as the analogy of the classical Fourier Transform and can be realized by the eigendecomposition of $\mathbf{L}$:
\begin{equation}
    \mathbf{L} = \mathbf{U} \operatorname{diag}(\boldsymbol{\lambda}) \mathbf{U}^T,
    \label{eq_GFT}
\end{equation}
where $\mathbf{U}$ is the orthonormal eigenvalue matrix and $\boldsymbol{\lambda}$ is the vector of eigenvalues. 
In GSP, the Laplacian matrix $\mathbf{L}$ of an undirected and unweighted is a symmetric semi-definite matrix. 
The eigenvectors serve as graph Fourier bases and eigenvalues represent graph frequencies.
The eigenvalue-eigenvector pairs are sorted in increasing order: smaller eigenvalues correspond to smoother variations (low frequencies) in the graph signal and larger eigenvalues correspond to rapid variations (high frequencies) \cite{Ortega_graph_2018}. 

Spectral operations can be conducted through applying filters $\sum_{f = 1}^{F}{h(\boldsymbol{\lambda})}_f$ to the signal $\boldsymbol{x}$ in the spectral domain through GFT:
\begin{equation}
    \tilde{\boldsymbol{x}} = \mathbf{U} \sum_{f = 1}^{F}{h(\boldsymbol{\lambda})}_f \mathbf{U}^T \boldsymbol{x},
    \label{eq_graph_conv}
\end{equation}
where $\tilde{\boldsymbol{x}}$ is the processed signal.
The formulation in \eqref{eq_graph_conv} is also known as the graph convolution. 
By implementing various filters, such as high-pass and low-pass filters, we can perform tasks like denoising and feature enhancement \cite{tremblay2018design}.

\subsection{LLM preliminaries}
The Generative Pre-trained Transformer 3 (GPT-3) is a transformer-based large-scale autoregressive language model developed by OpenAI. 
With 175 billion parameters, it is significantly larger than its predecessor, GPT-2, and was designed to excel in task-agnostic few-shot learning. 
GPT-3 can perform a wide array of NLP tasks, such as text generation, translation, summarization, and question answering, without requiring task-specific fine-tuning. Instead, it leverages in-context learning, where the model can adapt to new tasks by interpreting prompts and few-shot demonstrations directly within its input context \cite{NEURIPS2020_GPT3}.
The GPT-3 is capable perform various tasks, such as creative writing and generating code, highlighting its broad applicability \cite{floridi2020gpt}. 
Despite these strengths, GPT-3 has notable limitations, such as struggles with long-term coherence, logical consistency, and susceptibility to generating factually incorrect or biased content.
Building on the GPT-3 foundation, GPT-4 introduces several key advancements that address some of these limitations of GPT-3. 
GPT-4 offers improved performance across a wider range of tasks and demonstrates enhanced capabilities in reasoning, inference, and contextual understanding by incorporating improvements such as multimodal processing, larger context window, and optimized scaling laws to deliver more accurate and reliable outputs \cite{achiam2023gpt4}. 
These technical improvements enable GPT-4 to tackle more complex tasks, such as multimodal reasoning and handling intricate logical chains, with greater precision.
For example, the GPT-4 is capable of scoring in the top 10\% on simulated bar exams, compared to GPT-3.5, which could score only at the bottom 10\% \cite{bubeck2023sparks}. 

\section{Methodology}
\label{sec_methods}
\subsection{Methodology overview}
The LLM-OSR algorithm reconstructs missing graph signals by combining GSP-based processing and LLM prediction. 
To provide an intuition of the LLM-OSR, it first enhances signal features with GSP and then predicts missing values via the LLM for time-varying spatiotemporal data.
The entire process operates online, meaning that the spatiotemporal reconstruction is performed in real time as new signal observations are continuously received and processed by the LLM-OSR.
The signal observation model used in LLM-OSR can be expressed as
\begin{equation}
    \boldsymbol{o}[t] = \mathbf{M}(\boldsymbol{x}_g[t]+\boldsymbol{\epsilon}[t]),
\end{equation}
where $\mathbf{M}$ is the observation mask, $\boldsymbol{x}_g[t]$ is the ground truth graph signal, and $\boldsymbol{\epsilon}[t])$ is the i.i.d. zero-mean additive Gaussian noise to the graph signal. 
Consider a graph $\mathcal{G}$ with a total of $N$ nodes, and a subset of these nodes $\mathcal{O} \subseteq \mathcal{V}$, where only $O$ out of $N$ nodes are observed.
Using the observation set $\mathcal{O}$, we can construct a masking matrix $\mathbf{M}$ to model the signal observation:
\begin{equation}
    \mathbf{M} = \text{diag}([\mathcal{O}(v_1), \dots, \mathcal{O}(v_O)]^T),
\end{equation}
where the membership of a node $v_i$ in the observation set is defined as:
\begin{equation}
\mathcal{O}(v_i) =
\begin{cases} 
    1, & \text{if } v_i \in \mathcal{O}, \\
    0, & \text{if } v_i \notin \mathcal{O}.
\end{cases}
\end{equation}
Here, $\mathcal{O}(v_i)$ acts as an indicator function that determines whether $v_i$ belongs to the observed set, and $\mathbf{M}$ is a diagonal matrix with entries corresponding to these indicator values.
This assumption is fundamental in various applications, including climate change analysis \cite{Giraldo_2022_smooth}, skeleton-based gait recognition \cite{chen2024spatiotemporal}, and brain studies \cite{Bai_2023_Smoothness}.  
The workflow of one iteration of LLM-OSR is shown in Figure~\ref{fig_flow_overview}.
An overview of the LLM-OSR framework is shown in Algorithm~\ref{alg_llm_osr}.

\begin{algorithm}[htb]
\caption{LLM-OSR Overview}
\label{alg_llm_osr}
\begin{algorithmic}[1]
\State \textbf{Train Phase:}
\State Initialize and train the GSP-based spatial-temporal signal handler as seen in Algorithm~\ref{alg_GSP_handler}
\State \textbf{Testing Phase:}
\State Deploy the trained GSP-based spatial-temporal signal handler
\State Prepare the LLM-based spatial-temporal signal predictor
\While{new observations $\boldsymbol{o}[t]$ are available}
    \State Gather $\hat{\boldsymbol{x}}[t-1]$, the previous signal estimate
    \State Process the observations $\boldsymbol{o}[t]$ with the GSP-based 
    \Statex \hspace{12 pt} spatial-temporal signal handler and Collect the 
    \Statex \hspace{12 pt} processed observations $\tilde{\boldsymbol{o}}[t]$
    \State  Pass $\tilde{\boldsymbol{o}}[t]$ and $\hat{\boldsymbol{x}}[t-1]$ into the LLM-based spatial-
    \Statex \hspace{12 pt} temporal signal predictor
    \State Operate the LLM-based spatial-temporal signal 
    \Statex \hspace{12 pt} predictor as seen in Algorithm~\ref{alg_LLM_predictor}
    \State Collect reconstructed $\hat{\boldsymbol{x}}[t]$
\EndWhile
\end{algorithmic}
\end{algorithm}

\subsection{GSP-based Spatial-temporal Signal Handler}

\begin{figure*}[htb]
    \centering
    \includegraphics[trim={0 380 10 70},clip,width=\linewidth]{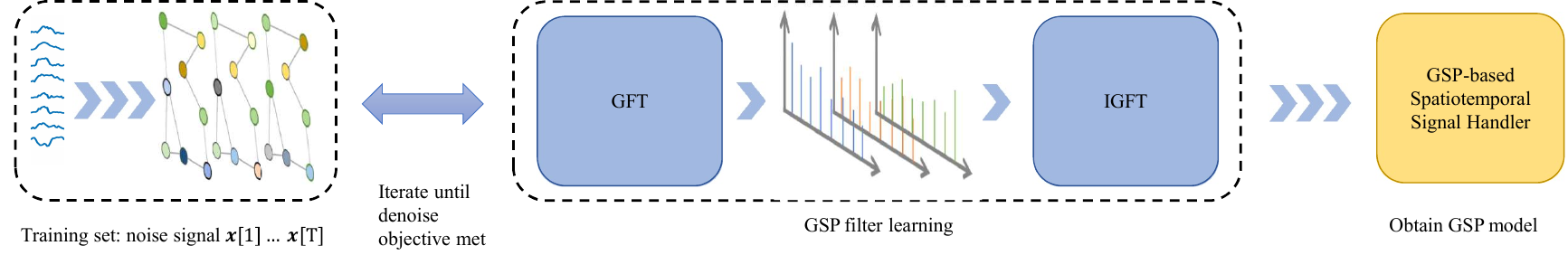}
    \caption{The training process of the GSP-based spatial-temporal signal handler.}
    \label{fig_GSP}
\end{figure*}

The proposed GSP-based Spatial-temporal Signal Handler aims to leverage GSP techniques to denoise and enhance spatial-temporal signals to prepare the data for the LLM-based spatial-temporal signal predictor. 

During the training phase of LLM-SRO, the goal is to learn the optimal filter parameters within the GSP-based spatial-temporal signal handler from the training data.
In our case, since we assume that the signal observations received by the LLM-OSR contain noise and missing nodes, we would like to let the GSP-based spatial-temporal handler perform a denoising task and feed the denoised observations to the LLM-based spatial-temporal signal predictor.
To give an overview, the LLM-OSR takes a universal approach to learning the filter parameters $\sum_{f = 1}^{F}{h(\boldsymbol{\lambda})}_f$ in \eqref{eq_graph_conv} by iteratively applying the graph convolution \eqref{eq_graph_conv} to the training data, calculating the loss of the estimation, and updating the filter parameter. 
We augment the training set by concatenating multiple copies to increase the number of time samples. During each iteration, the GSP-based spatial-temporal handler is trained using data from a single time instance.
First, we apply the graph convolution operation in \eqref{eq_graph_conv} to obtain the signal representation. 
Next, we compute the Mean Absolute Error (MAE) between the predicted output and the ground truth at each time instance $t$, serving as the performance metric for the current filter:
\begin{equation}
    \text{MAE}[t] = \frac{1}{N} \sum_{n = 1} ^ {N} |x_{g, n}[t]-\tilde{x}_n[t]|,
    \label{eq_MAE}
\end{equation}
where $\tilde{\boldsymbol{x}}$ is the processed graph signal, $\hat{x}_n[t]$ it the processed signal on the $n^{th}$ node within $\tilde{\boldsymbol{x}}$, and $x_{g, n}[t]$ is the signal on the $n^{th}$ node within the ground truth signal $\boldsymbol{x}_g[t]$.
The actual learning and updating of the filter are done by calculating the gradient of the MAE with respect to the filter parameters $\sum_{f = 1}^{F}{h(\boldsymbol{\lambda})}_f$, denoted as $\nabla_{\mathbf{H}} \text{MAE}$ then updated the filter parameters using the gradient descent rule:
\begin{equation}
    \sum_{f = 1}^{F}{h(\boldsymbol{\lambda})}_f = \sum_{f = 1}^{F}{h(\boldsymbol{\lambda})}_f - \eta \cdot \nabla_{h} \text{MAE}[t],
    \label{eq_filter_update}
\end{equation}
where $\eta$ is the learning rate. 
We repeat this process is repeated iteratively until we run out of training samples or when the MSE converges to a steady value (early stopping).
Since the graph filter parameters are applied within the graph convolution, it is essentially learned based on the knowledge of both the graph structure and associated graph signals. These parameters are optimized to minimize noise and preserve important spatial and spectral graph features in the data.

Once training is complete, the learned GSP filter parameters are fixed and stored in the GSP-based spatial-temporal signal handler for use during the test phase. 
We assume that the signals in the training set and the testing set have similar spectrums. 
During the test phase, the pre-trained GSP-based spatial-temporal signal handler is applied to unseen test samples $\boldsymbol{o}[t]$ as they are received in real-time to enhance their quality by denoising.
In other words, since each test sample $\boldsymbol{o}[t]$ is an observation with missing values and noise, the pre-trained GSP-based spatial-temporal signal handler is applying a series of graph filters $\sum_{f = 1}^{F}{h(\boldsymbol{\lambda})}_f$ to $\boldsymbol{o}[t]$  to denoise using the graph convolution \eqref{eq_graph_conv}.
Since the process is completed through the graph convolution \eqref{eq_graph_conv}, the signal observations are processed with the knowledge of the underlying spatial structure of the graph $\mathcal{G}$.
The logic of training and deploying the GSP-based spatial-temporal signal handler data can be found in Figure~\ref{fig_GSP}. 

\begin{algorithm}[htb]
\caption{GSP-based spatial-temporal signal handler}
\label{alg_GSP_handler}
\begin{algorithmic}[1]
\State \textbf{Train Phase:}
\State Given the training data $\boldsymbol{x}[1] \dots \boldsymbol{x}[T]$
\State Initialize graph Laplacian $\mathbf{L = D - A}$ \eqref{eq_laplacian} and GFT $\mathbf{L = U \Lambda U}^T$ \eqref{eq_GFT}
\State Initialize filter parameters $h(\mathbf{\Lambda)}$
\While{halting condition not met}
    \State Apply $\tilde{\boldsymbol{x}}[t] = \mathbf{U} \text{diag}(h(\mathbf{\Lambda})) \mathbf{U}^T \boldsymbol{x}[t]$, the graph 
    \Statex \hspace{12 pt} convolution, as seen in \eqref{eq_graph_conv}
    \State Compute MAE at each time instance $t$:
    \Statex \hspace{12 pt} MAE$[t] = \frac{1}{N} \sum_{n = 1}^{N} |x_{g, n}[t] - \tilde{x}_n[t]|$ as seen in \eqref{eq_MAE}
    \State Calculate gradient $\nabla_{h} \text{MAE}$ with respect to $h(\Lambda)$
    \State Update filter parameters using gradient descent:
    \Statex \hspace{12 pt} $h(\Lambda) = h(\Lambda) - \eta \cdot \nabla_{h} \text{MAE}$ as seen in \eqref{eq_filter_update}
\EndWhile
\State Export the parameters after training is complete
\State \textbf{Test Phase:}
\For{each received test observation sample $\boldsymbol{o}[t]$ containing noise and missing node observation}
    \State Process $\boldsymbol{o}[t]$ using graph convolution and the trained 
    \Statex \hspace{15pt}filter $\text{diag}(h(\Lambda))$: $\tilde{\boldsymbol{o}}[t] = \mathbf{U \text{diag}(h(\Lambda)) U}^T \boldsymbol{o}[t]$
    \State Apply $\tilde{\boldsymbol{o}}[t]$ to be fed into LLM for reconstruction
\EndFor
\end{algorithmic}
\end{algorithm}

\subsection{LLM-based spatial-temporal signal predictor}

\begin{figure*}[htb]
    \centering
    \includegraphics[trim={0 400 10 60},clip,width=\linewidth]{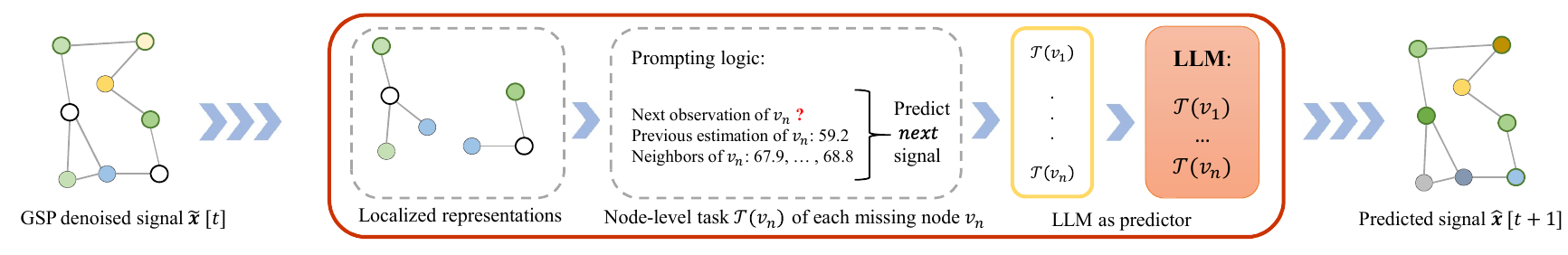}
    \caption{The LLM-based Spatial-temporal Signal Predictor.}
    \label{fig_LLM}
\end{figure*}

The first step of our LLM-based spatial-temporal signal predictor is a reverse embedding function to process the denoised signals. 
In NLP tasks, the embedding process typically involves transforming words, phrases, or sentences into numerical vectors by mapping them into a designated feature space \cite{qureshi2019eve}.
In conventional graph embedding tasks, data on the graph is typically transformed into a feature space using methods such as GCN or Node2Vec, similar to traditional NLP embeddings \cite{grover2016node2vec}. 
Here in the LLM-OSR, we take a reverse approach to the conventional embedding approaches such as the Node2Vec. 
Since the time-varying graph signal observations $\boldsymbol{o}[t]$ are already numerical, instead of embedding data into designated space, we directly represent the local topological connections of nodes on the graph in the spatial domain through LLM natural language expressions (English passages), which consists of text and numbers. 
Mathematically, extracting the 1-hop localized neighbors of the node $ v_i $ can be achieved by identifying the nonzero entries in the $ i^{\text{th}} $ row of the adjacency matrix $ \mathbf{A} $. This can be expressed as:
\begin{equation}
    \text{Neighbors}(v_i) = \mathcal{A}_{v_i} =\{ v_j \mid \mathbf{A}[i, j] \neq 0, \, j \neq i \},
    \label{eq_neighborhood}
\end{equation}
where $\text{Neighbors}(v_i)$ represents the set of neighboring nodes of $v_i$ .
Notice that in our problem setting, there is only a subset of the nodes being observed. 
So in the implementation of LLM-OSR, we only consider the processed node signals from the observed node neighbors. 

These expressions are then further put into the prompt as the tasks $\mathcal{T}(v_i)$ on each node to be solved by the LLMs, specifying the LLM to conduct a prediction based on the spatial-temporal information.
In the spatial domain, the task $\mathcal{T}(v_i)$ will consist of an expression of which nodes are neighbors to node $v_i$ and the processed observed values of its neighbors. 
In the temporal domain, the task $\mathcal{T}(v_i)$ will consist of an expression of the past estimated signal values of node $v_i$.
Conceptually, we can understand the task of the LLM as to aggregate neighbor signals of $v_i$ with self-aggregation; this task can be denoted as:
\begin{equation}
    \mathcal{T}(v_i) = \text{agg}\left( \{ (\hat{x}_i[t-1], \tilde{x}_j[t]) \mid j \in (\mathcal{A}_{v_i} \cup \mathcal{O}) \} \right).
    \label{eq_message}
\end{equation}
This approach shifts the role of the LLM from performing logical reasoning or mathematical calculations, which is a known weakness of the LLM, to generating data based on semantic understanding and context, leveraging the strengths of LLM in processing natural language descriptions.
The aim of the LLM-based Spatial-temporal Signal predictor is to let the LLM leverage the smoothness assumption to infer the target values of each node based on the values of its neighboring nodes (spatial) and the past estimation (temporal).

The LLMs used within our LLM-OSR framework are GPT-3.5-turbo and GPT-4-o mini, which belong to the GPT-3 and GPT-4 families, respectively. 
The GPT-3 leverages in-context learning to perform various tasks—such as translation, question answering, and text completion—without requiring task-specific fine-tuning \cite{NEURIPS2020_GPT3}. 
The architecture of the GPT-3 is based on transformers and was trained on a diverse dataset sourced primarily from the internet. 
GPT-4 further builds upon the GPT-3 foundation, incorporating key advancements such as multimodal capabilities, a larger context window, optimized scaling laws, and improved safety mechanisms \cite{achiam2023gpt4}. 
These enhancements result in significant improvements in inference and logical reasoning of GPT-4, demonstrating better performance compared to GPT-3 at understanding prompts, generating more coherent and contextually appropriate responses, and handling complex tasks requiring step-by-step reasoning \cite{bubeck2023sparks}.
These improvements make GPT-4 more suitable for our tasks of online spatial-temporal signal reconstruction, which involves complex reasoning and spatial-temporal pattern recognition.

The integration of LLMs with the rest of the components within the LLM-based Spatial-temporal Signal Predictor is facilitated through the OpenAI API.
To be specific, LLM-OSR directly utilizes the pre-trained model provided and deployed on the OpenAI server.
That is, the LLM-OSR does not perform any additional training or fine-tuning. 
The API provides access to the vanilla model as-is, leveraging its existing knowledge and capabilities to generate predictions. 
This allows us to take full advantage of the robust, generalized understanding embedded in the model, ensuring a streamlined and efficient process for spatial-temporal signal forecasting.

We followed the approach of structuring the interaction of the reversely embedded graph signal expression with the LLM using a dual-role setup where the LLM contains a system role and a user role. 
The system role serves as the supervision guide, providing the global task context and specific constraints to shape the behavior of the LLM. 
For our spatiotemporal task, this role defines the objective as predicting the current value of a graph node based on its previous value and the values of its neighbors, while also enforcing strict output requirements. 
In our LLM-OSR, the system instructs the LLM to produce only a single numeric value as output, rounded to a certain decimal place, and to avoid any extraneous text or reasoning. 
This ensures consistency and simplicity in the generated responses.
Below is an example of system role content we used in LLM-OSR:
\begin{quote}
\it{The spatiotemporal task is to predict the current number on a graph based on its previous value and the value of its neighbors.}
\end{quote}
The user role, on the other hand, dynamically generates task-specific prompts that supply the LLM with the required details for each individual prediction. 
Here is an example of directly expressing the neighborhood relationship along with the signals for the user role using  natural language expression:
\begin{quote}
\it{Each indexed content is independent. Make 1 numeric prediction per indexed context. Precision round to 1 decimal point. Do not output text. Do not recall memories. Time 1439, Entity index: 322. Previous: 61.5, Neighbors: [63.9, 57.4].}
\end{quote}
These prompts include precise temporal and spatial context, such as the time index, the node index, the previous value of the node, and a list of observed values from its neighbors. 
By structuring the user prompts this way, we ensure that each query to the LLM is both clear and contextually complete, reducing ambiguity in the response. 

Furthermore, the interaction process is optimized by batching multiple prompts for efficiency. 
When a new observation is received in the LLM-OSR, the user role creates a batch of prompts corresponding in real-time. 
These batched prompts are sent to the LLM in a single API call, with each prompt corresponding to the task $\mathcal{T}(v_i)$ on a single node, leveraging the prompt-response structure to efficiently handle multiple predictions.

Notice that we further included an error-checking function for invalid LLM responses.
If an LLM call fails, such as when the LLM response does not include a valid numeric value, our implementation automatically retries the failed predictions. 
This retry mechanism regenerates the prompts for the unresolved cases and resubmits them to the LLM, with a predefined maximum number of retries. 
This additional error-checking function ensures that exceptions and invalid LLM outputs are handled robustly, even in cases where the LLM struggles to produce a valid prediction.

\begin{figure*}[htb]
    \centering
    \includegraphics[trim={0 375 0 0},clip,width = \linewidth]{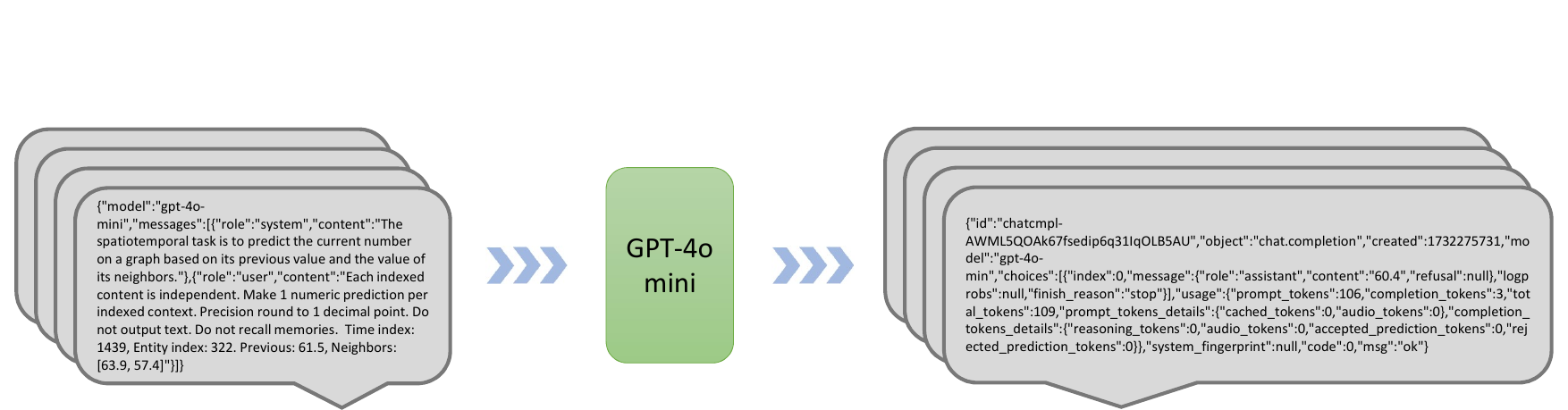}
    \caption{The prompts prepared for LLM and the responses generated by the LLM (GPT-4o mini).}
    \label{fig_prompts}
\end{figure*}

By combining the system and user roles with these efficient handling mechanisms, this structured approach allows the LLM to focus on its core strength of generating numeric predictions while providing the necessary temporal and spatial context for accurate, context-aware results.
An illustration of the prompts that provide the content and context to the LLM for solving the node-level tasks $\mathcal{T}(v_i) \big|_{v = 1 \dots N}$  can be found in Figure~\ref{fig_prompts}.

\begin{algorithm}[htb]
\caption{LLM-based spatial-temporal signal predictor}
\label{alg_LLM_predictor}
\begin{algorithmic}[1]
\State Initialize the LLM model and form the system and user role prompt templates
\While{new processed observations $\tilde{\boldsymbol{o}}[t]$ are available from the GSP-based spatial-temporal signal handler}
    \State Collect the output from the GSP-based filter 
    \Statex \hspace{12 pt} spatial-temporal signal handler
    \For{each missing node $v_i$ in set $\mathcal{V}-\mathcal{O}$}
        \State Collect $\hat{x}_i[t-1]$, the previous estimation of $v_i$
        \State Collect node neighborhood for $v_i$ using \eqref{eq_neighborhood}
        \Statex \hspace{27 pt} and the corresponding processed observed   
        \Statex \hspace{27 pt} neighbor signals of node $v_i$ in $\tilde{\boldsymbol{o}}[t]$
        \State Form LLM task $\mathcal{T}(v_i)$ from aggregation \eqref{eq_message} 
        \Statex \hspace{27 pt} and their corresponding prompts
        \State Feed $\mathcal{T}(v_i)$ and their corresponding prompts into \Statex \hspace{27 pt} the LLM 
        \Statex \hspace{27 pt} Let LLM predict $\hat{x}_i[t]$  
    \EndFor \hspace{1 pt} (Retry if output invalid)
    \State Collect all node reconstructions and map them to $\hat{\boldsymbol{x}}[t]$
\EndWhile
\end{algorithmic}
\end{algorithm}

\section{Experiments and Discussion}
\label{sec_experiments}
\subsection{Experiment Setting}
Here we will provide a brief discussion about the datasets, tested algorithms, and experiment settings.
\subsubsection{Dataset Description}
\begin{figure*}[htb]
    \centering
    \includegraphics[trim={40 310 40 310},clip,width = \linewidth]{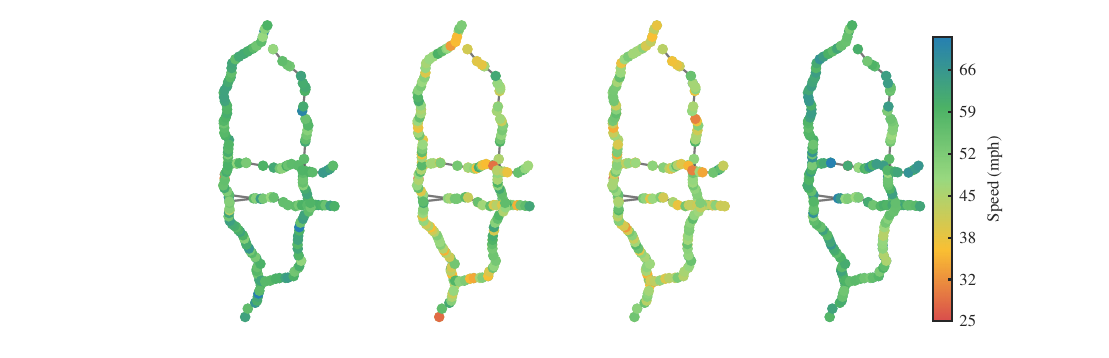}
    \caption{The Seattle loop dataset at 4 different time instances.}
    \label{fig_traffic}
\end{figure*}

\begin{itemize}
    \item \textbf{Traffic Data:} We utilize the publicly available Seattle Loop Detector Dataset \cite{seattle_loop_data}, which contains traffic flow data collected from loop detectors on the highways in the Seattle area. 
    This dataset provides hourly traffic readings that are essential for analyzing spatiotemporal patterns. 
    The experimental setup includes the addition of Gaussian noise with variances of 1.0 and 1.5 to evaluate the robustness of the models under varying levels of data corruption.
    The graph topology is constructed by mapping the physical locations of $N = 323$ loop detectors to their corresponding positions along the actual highway path.
    Each loop detector is a node on the graph $\mathcal{G}$.
    The traffic speed is recorded in 5-minute intervals. 
    We selected a sub-portion of the signal consisting of 7 days of reading, making the size of the data $\mathbb{R}^{323\times2016}$.
    The recordings from the first 576 will be in the training set to tune or learn the model parameters and the rest are in the testing set.
    An illustration of 4 different time instances of this time-varying dataset can be found in Figure~\ref{fig_traffic}, 
    \item \textbf{Meteorological Data:} Hourly wind speed and temperature data are obtained from NOAA \cite{noaa_weather_data}. 
    Because of the behavior differences between wind speed and temperature,  we analyze them as separate datasets in the experiments.
    Each node in the dataset corresponds to a geographic location defined by its latitude and longitude. 
    We selected $N = 197$ stations that contain no missing recordings in 3 consecutive days, giving us  $\mathbb{R}^{197\times96}$. 
    We split the first 24 time steps into the training set and the rest into the testing set.
    To capture the spatial dependencies among nodes, a k-nearest-neighbor (kNN) graph is constructed, where the edges' weights are computed by using a Gaussian kernel method which is described in the GNLMS framework \cite{yan2023graph}.
\end{itemize}
We set the node observation ratio to be 70$\%$ for all the datasets. 
The missing nodes are missing throughout the entire experiment, making it challenging to infer the missing signals without the utilization of the graph topology. 
The goal of the experiments is that given an observation $\boldsymbol{o}[t]$ that is only partially observed, reconstruct the ground truth signal $\boldsymbol{x}_g[t]$ from the observation and past $p$ estimations.

\subsubsection{Considered Algorithms}
We consider 2 distinct settings of the LLM-OSR.
The first setting is to use GPT-3.5 turbo as the LLM within the LLM-OSR; we denote this setting as the LLM-OSR-3.5. 
The second setting is to use GPT-4-o mini as the LLM within the LLM-OSR, denoted as LLM-OSR-4.
The 2 LLM-OSR variants will be evaluated against a variety of baseline algorithms, including graph adaptive filters, graph time-series analysis algorithms, and GNNs: 
\begin{itemize}
    \item \textbf{GLMS} \cite{Lorenzo_2016_LMS}: An adaptive filter designed for online graph signal estimation under Gaussian noise, derived from an LMS optimization problem.
    \item \textbf{GNLMS} \cite{Spelta_2020_NLMS}: A variant of the GLMS that incorporates spectral normalization to enhance performance.
    \item \textbf{GNS} \cite{peng2023adaptive}: An adaptive filter developed for online graph signal estimation under impulsive noise, derived from an $L_1$ optimization problem.
    \item \textbf{GCN} \cite{kipf2016semi}: A widely recognized graph neural network (GNN) where each layer applies the graph convolution \eqref{eq_graph_conv}, incorporating spatial normalization and a non-linear activation function.
    \item \textbf{GVARMA} \cite{Isufi2019gvarma}: A time-series analysis method that extends the classical VARMA model into the graph signal processing (GSP) domain by defining ARMA parameters using the graph Fourier transform (GFT) in the graph spectral domain.
    \item \textbf{GGARCH} \cite{Hong_GGARCH_2023}: A time-series analysis method analogous to the classical GARCH model, adapted to the GSP domain by defining GARCH parameters using the GFT in the graph spectral domain.
    \item \textbf{RGDAN} \cite{FAN_2024_RGDAN}: A GNN architecture that combines graph diffusion-based modeling with spatial and temporal attention mechanisms to capture complex relationships in graph-structured data.
\end{itemize}
Notice that other than the adaptive filters and the LLM-OSRs, the other algorithms are offline algorithms during training.

\subsubsection{Computational Environment}

The experiments are conducted on a workstation equipped with the following hardware and software:

\begin{itemize}
    \item \textbf{CPU:} Intel Core i9-13900K
    \item \textbf{GPU:} NVIDIA RTX 4090 with 24GB of G6X memory
    \item \textbf{Operating System:} Windows 11
\end{itemize}
These computational resources provide sufficient computational power for handling large-scale graph computations and training the LLM-OSR model.


\subsection{LLM-OSR on Traffic Prediction}
The enhanced multilingual proficiency of GPT-4 enables it to excel in low-resource languages, and its optimized scaling laws ensure predictable improvements. 
LLM on graphs perform well not because of leakage, LLM understands graphs as languages instead of the topological structures, and the node label (signal) is more important than structure: \cite{huang2023can_LLM_graph}.

The results of our experiments, as summarized in Tables \ref{table_traffic_RMSE} and \ref{table_traffic_MAE}, demonstrate the significant performance improvements achieved by the LLM-OSR models in traffic prediction tasks on the Seattle Loop Dataset. 
Specifically, LLM-OSR-4 consistently outperforms all baseline models in terms of both RMSE and MAE. These include GSP-based methods like GLMS, GNLMS, and GNS, GNN-based algorithms such as GCN and RGDAN, and time-series analysis algorithms like GVARMA and GGARCH.
These findings emphasize the ability of LLMs to understand graph-encoded traffic data by interpreting node-level signals as linguistic constructs instead of reading numerical data stored as graph representations or numerical embeddings. 
This unique approach allows LLM-OSR to extract richer, more contextually relevant patterns, thereby leading to superior predictive performance.

The experimental results further reveal a performance gap between LLM-OSR-3.5 and LLM-OSR-4. While GPT-4-based LLM-OSR-4 achieves good performance, GPT-3.5-based LLM-OSR-3.5 struggles significantly, often performing worse than many baseline graph-based methods. This performance disparity highlights the enhanced modeling capabilities of GPT-4, which can better understand the prompt and node-level signals and adapt to noisy inputs compared to GPT-3.5. It also shows that earlier LLM versions like GPT-3.5 may lack the robustness required for spatiotemporal graph tasks.

\begin{table}[htb]
\centering
\caption{Experiment RMSE for Seattle Loop Dataset}
\begin{tabular}{ccc}
    \toprule
    Model & 1.0 & 1.5 \\
    \midrule
    LLM-OSR-3.5 & 12.23 ± 3.9e+00 & 13.76 ± 4.0e-00\\
    LLM-OSR-4 & \textbf{4.05 ± 1.6e-01} & \textbf{4.69 ± 6.8e-03}\\
    GLMS & 8.04 ± 2.1e-03  & 8.07 ± 2.1e-03\\
    GNLMS & 7.91 ± 1.2e-03 & 7.91 ± 9.0e-04\\
    GNS & 8.55 ± 1.5e-03 & 8.55 ± 2.3e-03\\
    GCN & 26.93 ± 7e-02 & 26.95 ± 4e-02\\
    GVARMA & 21.75 ± 1.3e-03 & 21.75 ± 1.3e-03 \\
    GGARCH & 21.73 ± 4.0e-04 & 21.73 ± 1.3e-03 \\
    RGDAN & \textit{5.32 ± 3.2e-01} & \textit{6.61 ± 2.1e+00}\\
    \bottomrule
\end{tabular}
\label{table_traffic_RMSE}
\end{table}

\begin{table}[htb]
\centering
\caption{Experiment MAE for Seattle Loop Dataset}
\begin{tabular}{ccc}
    \toprule
        Model & 1.0 & 1.5 \\
    \midrule
    LLM-OSR-3.5 & 3.52 ± 3.8e-01 & 4.23 ± 3.3e-01\\
    LLM-OSR-4 & \textbf{2.88 ± 2.3e-02}& \textbf{3.62 ± 7.4e-03}\\
    GLMS & 5.09 ± 1.8e-03 & 5.12 ± 2.4e-03\\
    GNLMS & 4.88 ± 9.3e-04 & 4.89 ± 7.6e-04\\
    GNS & 4.69 ± 1.6e-03 & 4.70 ± 1.2e-03\\
    GCN & 19.22 ± e+00& 19.22 ± e+00\\
    GVARMA & 18.53 ± 3.7e-04 & 18.53 ± 1.2e-03 \\
    GGARCH & 18.54 ± 3.5e-04 & 18.55 ± 1.2e-03 \\
    RGDAN & \textit{3.23 ± 1.3e-01} & \textit{3.96 ± 1.2e+00}\\
    \bottomrule
    \end{tabular}
\label{table_traffic_MAE}
\end{table}

\subsection{LLM-OSR on Weather Prediction}

The performance of LLM-OSR models on weather prediction tasks, as presented in Tables \ref{table_wind_RMSE}, \ref{table_wind_MAE}, \ref{table_temp_RMSE}, and \ref{table_temp_MAE}, highlights their capability in handling spatiotemporal graph data under varying noise conditions. Gaussian noise with variances of 0.2, 0.6, and 1.0 was added to simulate real-world data. The results show that while LLM-OSR-4 excels under lower noise levels, its performance degrades more significantly compared to other models as noise variance increases, which reveals a limitation of current LLM-based approaches. 

For hourly wind speed prediction, LLM-OSR-4 achieves exceptional results. When the noise variance is 0.2 and 0.6 it outperforms all baselines. 
For noise variances of 0.6 and 1.0, LLM-OSR-4 ranks as the best-performing in terms of RMSE and second-best-performing in terms of MAE. 
Its second-best-performing cases are surpassed only by RGDAN, a recently proposed, sophisticated model that combines spatial and temporal embeddings. 
RGDAN achieves this by leveraging GNN diffusion attention mechanisms for spatial embeddings and with temporal attention for temporal embeddings. 
These embeddings are then integrated using transformer attention, making the RDGAN a powerful algorithm \cite{FAN_2024_RGDAN}.
it is worth noting that both RDGAN and GPT-4-o mini are attention-based algorithms, which highlights the potential for LLM-OSR to further enhance its performance.
We will discuss several potential approaches that could potentially boost the performance of LLM-OSR in the next section.
In hourly temperature prediction tasks, LLM-OSR-4 maintains its leading performance under low noise conditions for a noise variance of 0.2.
However, its performance declines as noise increases compared with RGDAN, but is still the second-best-performing algorithm among all the tested algorithms.
Looking at the results within the LLMs, the LLM-OSR-4 outperforms LLM-OSR-3.5 again, suggesting that the GPT-4-o mini is a more appropriate choice than the GPT-3.5 turbo for LLM-OSR.

The results suggest that the performance of LLM-OSR-4 will degrade as the noise variance increases in comparison to non-LLM methods such as RGDAN.
Theoretically, algorithms are expected to perform worse as noise increases, but LLM-OSR-4 appears more sensitive to this degradation. 
This is likely due to the inherent limitations of LLMs. 
While the GSP-based spatial-temporal handler is capable of noise reduction, it does not completely eliminate noise.
The residual noise within the processed signals $\tilde{\boldsymbol{o}}[t]$ leads to degraded predictions of LLM-OSR. 
Another potential factor that leads to reduced robustness under high noise conditions is the lack of fine-tuning or retraining of the LLM in the LLM-OSR. 
We acknowledge the limitations and have considered them in our experimental design. 
A more detailed discussion of this issue will be presented in the next section. 
Despite these challenges, the strong performance of LLM-OSR-4 under Gaussian noise conditions demonstrates its potential for spatial-temporal signal prediction tasks with noisy real-world datasets.

\begin{table}[htb]
\centering
\caption{Experiment RMSE for Hourly Wind Speed Prediction}
\begin{tabular}{cccc}
\toprule
Model & 0.2 & 0.6 & 1.0 \\
\midrule
LLM-OSR-3.5 & 4.58 ± 9.1e-01 & 5.11 ± 9.8e-01 & 5.06 ± 4.9e-01\\
LLM-OSR-4 & \textbf{1.39 ± 6.1e-03} & \textbf{1.70 ± 7.5e-03} & \textbf{1.91 ± 1.1e-02}\\
GLMS & 2.19 ± 1.8e-03 & 2.21 ± 2.9e-03 & 2.22 ± 5.4e-03\\
GNLMS & 2.18 ± 1.5e-03 & 2.20 ± 3.3e-03 & 2.21 ± 5.8e-03\\
GNS & 2.38 ± 5.8e-03 & 2.39 ± 6.7e-03 & 2.39 ± 8.2e-03\\
GCN & 2.65 ± 1e-02 & 2.82 ± 4.2e-01 & 2.68 ± 5e-02\\
RGDAN & \textit{1.83 ± 9.0e-02} & \textit{1.89 ± 5.2e-02} & \textit{1.97 ± 5.4e-02}\\
GVARMA & 4.07 ± 6.7e-03 & 4.09 ± 1.0e-02 & 4.11 ± 1.5e-02 \\
GGARCH & 3.71± 1.9e-03 &3.73 ± 3.2e-03 & 3.76 ± 6.7e-03 \\

\bottomrule
\end{tabular}
\label{table_wind_RMSE}
\end{table}

\begin{table}[htb]
\centering
\caption{Experiment MAE for Hourly Wind Speed Prediction}
\begin{tabular}{cccc}
\toprule
Model & 0.2 & 0.6 & 1.0 \\
\midrule
LLM-OSR-3.5 & 2.77 ± 8.8e-01 & 3.09 ± 8.4e-01 & 5.06 ± 4.9e-01\\
LLM-OSR-4 & \textbf{1.01 ± 6.3e-03} & \textbf{1.35 ± 6.3e-03} & \textit{1.54 ± 1.0e-02}\\
GLMS & 1.75 ± 1.5e-03 & 1.76 ± 1.9e-03 & 1.77 ± 3.7e-03\\
GNLMS & 1.74 ± 1.3e-03 & 1.75 ± 2.1e-03 & 1.76 ± 3.8e-03\\
GNS & 1.87 ± 3.8e-03 & 1.88 ± 4.5e-03 & 3.53 ± 5.9e-03\\
GCN & 2.10 ± 4.1e-02 & 2.22 ± 1.5e-02 & 2.12 ± 3.4e-02\\
RGDAN & \textit{1.35 ± 6.2e-02}& \textit{1.43 ± 3.5e-02} & \textbf{1.51 ± 3.6e-02}\\
GVARMA & 3.15 ± 4.2e-03& 3.17 ±  6.5e-03& 3.19 ± 9.0e-03\\
GGARCH & 3.10 ± 1.9e-03& 3.12 ± 3.9e-03& 3.14 ± 7.7e-03\\
\bottomrule
\end{tabular}
\label{table_wind_MAE}
\end{table}

\begin{table}[htb]
\centering
\caption{Experiment RMSE for Hourly Temperature Prediction}
\begin{tabular}{cccc}
\toprule
Model & 0.2 & 0.6 & 1.0 \\
\midrule
LLM-OSR-3.5 & 4.90 ± 9.2e-01 & 4.97 ± 6.8e-01 & 4.65 ± 7.7e-01\\
LLM-OSR-4 & \textbf{1.23 ± 7.5e-03} & \textit{1.54 ± 4.8e-03} & \textit{1.72 ± 6.0e-03}\\
GLMS & 4.41 ± 5.1e-03 &  4.43 ± 7.1e-03& 4.45 ± 9.1e-03 \\
GNLMS & 4.40 ± 5.7e-03 &  4.41 ± 7.0e-03 & 4.42 ± 9.1e-03 \\
GNS & 5.53 ± 1.6e-02 &  5.54 ± 2.9e-02 & 5.57 ± 1.8e-02 \\
GCN & 3.09 ± 5.3e-02 & 3.10 ± 6.3e-02 & 3.13 ± 7.4e-02\\
RGDAN & \textit{1.33 ± 1.9e-01} & \textbf{1.33 ± 9.3e-02} & \textbf{1.37 ± 1.3e-01}\\
GVARMA & 3.13 ±  2.7e-03& 3.17 ± 4.7e-03& 3.22 ± 4.4e-03\\
GGARCH & 3.22 ±  2.7e-03& 3.26 ± 4.7e-03& 3.30 ± 6.1e-03\\

\bottomrule
\end{tabular}
\label{table_temp_RMSE}
\end{table}

\begin{table}[htb]
\centering
\caption{Experiment MAE for Hourly Temperature Prediction}
\begin{tabular}{cccc}
\toprule
Model & 0.2 & 0.6 & 1.0 \\
\midrule
LLM-OSR-3.5 & 1.78 ± 1.3e-01 & 1.97 ± 8.0e-02 & 2.02 ± 8.2e-02\\
LLM-OSR-4 & \textbf{0.90 ± 4.1e-03} & \textit{1.19 ± 3.2e-03} & \textit{1.35 ± 3.6e-03}\\
GLMS & 3.03 ± 1.3e-03 &  3.05 ± 3.6e-03 & 3.07 ± 4.0e-03\\
GNLMS & 3.02 ± 1.6e-03 &  3.03 ± 3.1e-03 & 3.04 ± 3.4e-03\\
GNS & 3.58 ± 3.8e-03 & 3.55 ± 8.2e-03 & 3.53 ± 5.9e-03\\
GCN & 2.32 ± 4.1e-02 & 2.32 ± 1.5e-02 & 2.29 ± 4.4e-02 \\
RGDAN & \textit{0.98 ± 1.4e-01} & \textbf{0.99 ± 6.5e-02} & \textbf{1.04 ± 1.0e-01}\\
GVARMA & 2.29 ± 3.1e-03& 2.34 ± 5.3e-03& 2.39 ± 3.7e-03\\
GGARCH & 2.38 ± 2.8e-03& 2.43 ± 4.6e-03& 2.47 ± 4.5e-03\\

\bottomrule
\end{tabular}
\label{table_temp_MAE}
\end{table}

\section{Limitations and Future Work}
\label{sec_limitation}
The LLM-OSR demonstrated an impressive ability to capture complex relationships and patterns inherent in graph structures while conducting 1-step online prediction on time-varying graph signals. 
However, as we developed the LLM-OSR, several limitations and challenges emerged that must be addressed when leveraging LLMs on time-varying graph signals.

Let us discuss the LLM-related limitations in the LLM-OSR.
First, LLMs are known to have difficulties in understanding numerical data \cite{li2024simac}. 
In our experiments, there are rare occasions that the LLM will output a NaN instead of giving us a numeric output. 
This limitation can be addressed in the future when more powerful LLMs are proposed.
Diving deeper into the aspect of the intrinsic limitation of LLMs, we noticed that poorly designed prompts often fail to generate accurate numerical outputs or even any prediction at all. Prompts that are inaccurate or of low accuracy can significantly impair the capabilities of LLM. \cite{jang2023can}
To improve numerical understanding, prompts should be carefully designed to provide clear instructions and context\cite{jia2024gpt4mts}.
In LLM-OSR, LLMs are used as predictors, which means that we do not train or tune the LLMs and let LLMs make zero-shot predictions. 
The decision not to fine-tune the LLM-OSR models stems from several considerations.
However, fine-tuning large language models, such as GPT-4, requires significant computational resources and time. 
In addition, fine-tuning requires an appropriate balance between the size of the training dataset and the number of parameters in the model. 
In our case, the available datasets are relatively small compared to the parameter size of GPT-4, making fine-tuning or retraining the LLMs only marginally effective.
One potential workaround to retraining and fine-tuning the LLMs would be including examples within the prompt that demonstrate how numerical outputs are expected; this can transform the LLM predictors from zero-shot learners to few-shot learners, which is expected to help guide the model for more accurate predictions.
We expect there will be a performance increase if the LLMs are fine-tuned or deployed as a few-shot learning predictor instead of the current zero-shot predictor \cite{Chen_2023_zero_shot}.

There are also spatio-temporal graph-related limitations that could be addressed in future research.
In the spatial domain, our current LLM-OSR is limited to processing one node neighborhood per LLM prompt. 
When a single prompt is used to process multiple nodes, the outputs are frequently incomplete or contain extraneous elements, as LLMs exhibit limited capability in handling multiple sequences of numbers in a single call. 
In other words, whenever we attempt to process $N$ nodes together, LLM often returns a number of prediction outputs that are not $N$, and it becomes nearly impossible to align the output when the number of inputs and outputs are mismatched.
In the temporal domain, LLMs struggle to comprehend long temporal behaviors and predict longer temporal sequences\cite{maharana2024evaluating}.
These limitations could potentially be solved by more advanced graph representation approaches and advanced programming techniques.

During the development of LLM-OSR, we have encountered challenges in terms of scalability and computational complexity. 
Currently, the speed of completing each LLM call is constrained by the speed at which LLMs process the input tokens and generate output tokens. 
Combined with the fact that we are processing one node neighborhood per prompt, the run speed of LLM-OSR is significantly dragged down.
We also notice that in our current setting, each LLM call will have to provide content in the prompt describing the task along with the reversely embedded graph signals.
Otherwise, the performance will significantly decrease as we progress.
This is likely due to the fact that LLMs have limited long-term memory capabilities \cite{shahriar2024putting}. 
Similarly, when the LLM generates outputs, the numerical are often associated with contextual text even when the prompt asks it to output only numerical outputs without text. 
These limitations lead to a significant amount of unintended token usage that bottlenecks the I/O and bandwidth of LLM-OSR, making it challenging to scale to larger graphs and longer temporal sequences.
Other than improving the LLMs themselves, an approach that we plan to take in the future is to implement LLM-OSR using distributive techniques. 

Lastly, we would like to expand the fields of applications of LLM-OSR.
For instance, we aim to investigate how LLM-OSR performs in applications involving impulsive and heavy-tailed noise, such as communication systems \cite{karakucs2020modelling} and medical imaging \cite{lee2023deep}. A potential approach involves modeling the noise not with a Gaussian distribution, but with $\alpha$-stable distributions. 
These distributions are well-suited for such scenarios due to their heavy tails and impulsive characteristics. 
Incorporating $\alpha$-stable distributions into LLM-OSR may enhance its robustness by enabling it to handle extreme values more effectively, thereby improving model stability in datasets with long-tailed distributions \cite{herranz2004alpha, kuruoglu1997new}.
Furthermore, we aim to enhance the contextual handling capabilities of LLM-OSR to broaden its applications by leveraging its contextual inference power of LLM beyond numerical multivariate data. 
This expansion could pave the way for developing LLM-OSR variants tailored to CV and artificial intelligence applications in scientific domains, such as document image understanding \cite{kuruoglu2010using} and material science \cite{saffarimiandoab2021insights}.

\section{Conclusion}
\label{sec_conclusion}
The LLM-OSR algorithm shows significant potential in reconstructing spatial-temporal graph signals by combining GSP-based denoise handler with LLM-based prediction. Experimental results highlight the superior performance of LLM-OSR-4 in capturing spatial-temporal dependencies and it achieves high accuracy in signal reconstruction for traffic and weather datasets. 

While the current performance of LLM-OSR is promising, significant work remains to fully unleash the capabilities of LLMs in spatial-temporal prediction to address the current limitations.
The LLM-OSR could serve as a foundation to spark future studies, driving innovation and exploration in the intersection of large language models and dynamic graph signal prediction.


\section*{Acknowledgment}
This work is supported by the Tsinghua Shenzhen International Graduate School Start-up fund under Grant QD2022024C, Shenzhen Science and Technology Innovation Commission under Grant JCYJ20220530143002005, and Shenzhen Ubiquitous Data Enabling Key Lab under Grant ZDSYS20220527171406015.

\bibliographystyle{IEEEbib}
\bibliography{paper}

\begin{thebibliography}{10}

\bibitem{acosta2022multimodal}
J.~N. Acosta, G.~J. Falcone, P.~Rajpurkar, and E.~J. Topol,
\newblock ``Multimodal biomedical ai,''
\newblock {\em Nature Medicine}, vol. 28, no. 9, pp. 1773--1784, 2022.

\bibitem{sonkavde2023forecasting}
G.~Sonkavde, D.~S. Dharrao, A.~M. Bongale, S.~T. Deokate, D.~Doreswamy, and S.~K. Bhat,
\newblock ``Forecasting stock market prices using machine learning and deep learning models: A systematic review, performance analysis and discussion of implications,''
\newblock {\em International Journal of Financial Studies}, vol. 11, no. 3, pp. 94, 2023.

\bibitem{radford2019better}
A.~Radford, J.~Wu, D.~Amodei, D.~Amodei, J.~Clark, M.~Brundage, and I.~Sutskever,
\newblock ``Better language models and their implications,''
\newblock {\em OpenAI blog}, vol. 1, no. 2, 2019.

\bibitem{devlin2018bert}
J.~Devlin,
\newblock ``Bert: Pre-training of deep bidirectional transformers for language understanding,''
\newblock {\em arXiv preprint arXiv:1810.04805}, 2018.

\bibitem{floridi2020gpt}
L.~Floridi and M.~Chiriatti,
\newblock ``Gpt-3: Its nature, scope, limits, and consequences,''
\newblock {\em Minds and Machines}, vol. 30, pp. 681--694, 2020.

\bibitem{waisberg2023gpt}
E.~Waisberg, J.~Ong, M.~Masalkhi, S.~A. Kamran, N.~Zaman, P.~Sarker, A.~G. Lee, and A.~Tavakkoli,
\newblock ``Gpt-4: a new era of artificial intelligence in medicine,''
\newblock {\em Irish Journal of Medical Science (1971-)}, vol. 192, no. 6, pp. 3197--3200, 2023.

\bibitem{sun2021ernie}
Y.~Sun, S.~Wang, S.~Feng, S.~Ding, C.~Pang, J.~Shang, J.~Liu, X.~Chen, Y.~Zhao, Y.~Lu, et~al.,
\newblock ``Ernie 3.0: Large-scale knowledge enhanced pre-training for language understanding and generation,''
\newblock {\em arXiv preprint arXiv:2107.02137}, 2021.

\bibitem{chen2024application}
J.~Chen, S.~Li, Q.~Huang, S.~Yan, Z.~Xie, and Y.~Lu,
\newblock ``Application of kimi intelligent assistant in the teaching of water pollution control engineering course,''
\newblock {\em International Journal of Education and Humanities}, vol. 13, no. 3, pp. 39--43, 2024.

\bibitem{sohail2023decoding}
S.~S. Sohail, F.~Farhat, Y.~Himeur, M.~Nadeem, D.~{\O}. Madsen, Y.~Singh, S.~Atalla, and W.~Mansoor,
\newblock ``Decoding chatgpt: a taxonomy of existing research, current challenges, and possible future directions,''
\newblock {\em Journal of King Saud University-Computer and Information Sciences}, p. 101675, 2023.

\bibitem{navarro2023exploring}
A.~Navarro and F.~Casacuberta,
\newblock ``Exploring multilingual pretrained machine translation models for interactive translation,''
\newblock in {\em Proceedings of Machine Translation Summit XIX, Vol. 2: Users Track}, 2023, pp. 132--142.

\bibitem{deng2023llms}
X.~Deng, V.~Bashlovkina, F.~Han, S.~Baumgartner, and M.~Bendersky,
\newblock ``Llms to the moon? reddit market sentiment analysis with large language models,''
\newblock in {\em Companion Proceedings of the ACM Web Conference 2023}, 2023, pp. 1014--1019.

\bibitem{tang2023evaluating}
L.~Tang, Z.~Sun, B.~Idnay, J.~G. Nestor, A.~Soroush, P.~A. Elias, Z.~Xu, Y.~Ding, G.~Durrett, J.~F. Rousseau, et~al.,
\newblock ``Evaluating large language models on medical evidence summarization,''
\newblock {\em NPJ digital medicine}, vol. 6, no. 1, pp. 158, 2023.

\bibitem{miraglia2022brain}
F.~Miraglia, F.~Vecchio, C.~Pappalettera, L.~Nucci, M.~Cotelli, E.~Judica, F.~Ferreri, and P.~M. Rossini,
\newblock ``Brain connectivity and graph theory analysis in alzheimer’s and parkinson’s disease: the contribution of electrophysiological techniques,''
\newblock {\em Brain Sciences}, vol. 12, no. 3, pp. 402, 2022.

\bibitem{2024_Market_Qin}
D.~Qin and E.~E. Kuruoglu,
\newblock ``Graph learning based financial market crash identification and prediction,''
\newblock in {\em IEEE CAI}, 2024.

\bibitem{rostami2023community}
M.~Rostami, M.~Oussalah, K.~Berahmand, and V.~Farrahi,
\newblock ``Community detection algorithms in healthcare applications: a systematic review,''
\newblock {\em IEEE Access}, vol. 11, pp. 30247--30272, 2023.

\bibitem{yan_2023_BiSCNN}
Y.~Yan and E.~E. Kuruoglu,
\newblock ``Binarized simplicial convolutional neural networks,''
\newblock {\em Neural Networks}, 2024.

\bibitem{FAN_2024_RGDAN}
J.~Fan, W.~Weng, H.~Tian, H.~Wu, F.~Zhu, and J.~Wu,
\newblock ``{RGDAN}: A random graph diffusion attention network for traffic prediction,''
\newblock {\em Neural Networks}, vol. 172, pp. 106093, 2024.

\bibitem{xu2024quantum}
S.~Xu, F.~Wilhelm-Mauch, and W.~Maass,
\newblock ``Quantum feature embeddings for graph neural networks.,''
\newblock in {\em HICSS}, 2024, pp. 7633--7642.

\bibitem{bevans2023monitoring}
B.~Bevans, A.~Ramalho, Z.~Smoqi, A.~Gaikwad, T.~G. Santos, P.~Rao, and J.~Oliveira,
\newblock ``Monitoring and flaw detection during wire-based directed energy deposition using in-situ acoustic sensing and wavelet graph signal analysis,''
\newblock {\em Materials \& Design}, vol. 225, pp. 111480, 2023.

\bibitem{sharma2024emerging}
R.~Sharma and H.~K. Meena,
\newblock ``Emerging trends in eeg signal processing: A systematic review,''
\newblock {\em SN Computer Science}, vol. 5, no. 4, pp. 1--14, 2024.

\bibitem{Dong_Graph_ML_2020}
X.~Dong, D.~Thanou, L.~Toni, M.~Bronstein, and P.~Frossard,
\newblock ``Graph signal processing for machine learning: A review and new perspectives,''
\newblock {\em IEEE Signal Processing Magazine}, vol. 37, no. 6, pp. 117--127, 2020.

\bibitem{kipf2016semi}
T.~N. Kipf and M.~Welling,
\newblock ``Semi-supervised classification with graph convolutional networks,''
\newblock {\em ICLR}, 2017.

\bibitem{yu2017spatio}
B.~Yu, H.~Yin, and Z.~Zhu,
\newblock ``Spatio-temporal graph convolutional networks: A deep learning framework for traffic forecasting,''
\newblock {\em IJCAI}, 2018.

\bibitem{wang2022robust}
W.~Wang and Q.~Sun,
\newblock ``Robust adaptive estimation of graph signals based on welsch loss,''
\newblock {\em Symmetry}, vol. 14, no. 2, pp. 426, 2022.

\bibitem{yan_2022_sign}
Y.~Yan, E.~E. Kuruoglu, and M.~A. Altinkaya,
\newblock ``Adaptive sign algorithm for graph signal processing,''
\newblock {\em Signal Processing}, vol. 200, pp. 108662, 2022.

\bibitem{jin2024large}
B.~Jin, G.~Liu, C.~Han, M.~Jiang, H.~Ji, and J.~Han,
\newblock ``Large language models on graphs: A comprehensive survey,''
\newblock {\em IEEE Transactions on Knowledge and Data Engineering}, 2024.

\bibitem{huang2023can_LLM_graph}
J.~Huang, X.~Zhang, Q.~Mei, and J.~Ma,
\newblock ``Can llms effectively leverage graph structural information: when and why,''
\newblock {\em arXiv preprint arXiv:2309.16595}, 2023.

\bibitem{ye2023natural}
R.~Ye, C.~Zhang, R.~Wang, S.~Xu, Y.~Zhang, et~al.,
\newblock ``Natural language is all a graph needs,''
\newblock {\em arXiv preprint arXiv:2308.07134}, vol. 4, no. 5, pp. 7, 2023.

\bibitem{Ortega_graph_2018}
A.~Ortega, P.~Frossard, J.~Kovačević, J.~M.~F. Moura, and P.~Vandergheynst,
\newblock ``Graph signal processing: Overview, challenges, and applications,''
\newblock {\em Proceedings of the IEEE}, vol. 106, no. 5, pp. 808--828, 2018.

\bibitem{tremblay2018design}
N.~Tremblay, P.~Gon{\c{c}}alves, and P.~Borgnat,
\newblock ``Design of graph filters and filterbanks,''
\newblock in {\em Cooperative and Graph Signal Processing}, pp. 299--324. Elsevier, 2018.

\bibitem{NEURIPS2020_GPT3}
T.~Brown, B.~Mann, N.~Ryder, M.~Subbiah, J.~D. Kaplan, P.~Dhariwal, A.~Neelakantan, P.~Shyam, G.~Sastry, A.~Askell, S.~Agarwal, A.~Herbert-Voss, G.~Krueger, T.~Henighan, R.~Child, A.~Ramesh, D.~Ziegler, J.~Wu, C.~Winter, C.~Hesse, M.~Chen, E.~Sigler, M.~Litwin, S.~Gray, B.~Chess, J.~Clark, C.~Berner, S.~McCandlish, A.~Radford, I.~Sutskever, and D.~Amodei,
\newblock ``Language models are few-shot learners,''
\newblock in {\em Advances in Neural Information Processing Systems}, H.~Larochelle, M.~Ranzato, R.~Hadsell, M.~Balcan, and H.~Lin, Eds. 2020, vol.~33, pp. 1877--1901, Curran Associates, Inc.

\bibitem{achiam2023gpt4}
J.~Achiam, S.~Adler, S.~Agarwal, L.~Ahmad, I.~Akkaya, F.~L. Aleman, D.~Almeida, J.~Altenschmidt, S.~Altman, S.~Anadkat, et~al.,
\newblock ``Gpt-4 technical report,''
\newblock {\em arXiv preprint arXiv:2303.08774}, 2023.

\bibitem{bubeck2023sparks}
S.~Bubeck, V.~Chandrasekaran, R.~Eldan, J.~Gehrke, E.~Horvitz, E.~Kamar, P.~Lee, Y.~T. Lee, Y.~Li, S.~Lundberg, et~al.,
\newblock ``Sparks of artificial general intelligence: Early experiments with gpt-4,''
\newblock {\em arXiv preprint arXiv:2303.12712}, 2023.

\bibitem{Giraldo_2022_smooth}
J.~H. Giraldo, A.~Mahmood, B.~Garcia-Garcia, D.~Thanou, and T.~Bouwmans,
\newblock ``Reconstruction of time-varying graph signals via sobolev smoothness,''
\newblock {\em IEEE Transactions on Signal and Information Processing over Networks}, vol. 8, pp. 201--214, 2022.

\bibitem{chen2024spatiotemporal}
G.~Chen, X.~Chen, C.~Zheng, J.~Wang, X.~Liu, and Y.~Han,
\newblock ``Spatiotemporal smoothing aggregation enhanced multi-scale residual deep graph convolutional networks for skeleton-based gait recognition,''
\newblock {\em Applied Intelligence}, pp. 1--21, 2024.

\bibitem{Bai_2023_Smoothness}
W.~Bai,
\newblock ``Smoothness harmonic: A graph-based approach to reveal spatiotemporal patterns of cortical dynamics in fmri data,''
\newblock {\em Applied Sciences}, vol. 13, no. 12, 2023.

\bibitem{qureshi2019eve}
M.~A. Qureshi and D.~Greene,
\newblock ``Eve: explainable vector based embedding technique using wikipedia,''
\newblock {\em Journal of Intelligent Information Systems}, vol. 53, pp. 137--165, 2019.

\bibitem{grover2016node2vec}
A.~Grover and J.~Leskovec,
\newblock ``node2vec: Scalable feature learning for networks,''
\newblock in {\em SIGKDD}, 2016, pp. 855--864.

\bibitem{seattle_loop_data}
C.~of~Seattle,
\newblock ``Seattle loop detector data,'' \url{https://github.com/zhiyongc/Seattle-Loop-Data}, 2020.

\bibitem{noaa_weather_data}
{National Oceanic and Atmospheric Administration},
\newblock ``National oceanic and atmospheric administration (noaa) weather data,'' \url{https://www.noaa.gov/}, 2024.

\bibitem{yan2023graph}
Y.~Yan, R.~Adel, and E.~E. Kuruoglu,
\newblock ``Graph normalized-lmp algorithm for signal estimation under impulsive noise,''
\newblock {\em Journal of Signal Processing Systems}, vol. 95, no. 1, pp. 25--36, 2023.

\bibitem{Lorenzo_2016_LMS}
P.~D.~Lorenzo, S.~Barbarossa, P.~Banelli, and S.~Sardellitti,
\newblock ``Adaptive least mean squares estimation of graph signals,''
\newblock {\em IEEE Transactions on Signal and Information Processing over Networks.}, vol. 2, no. 4, pp. 555 -- 568, 2016.

\bibitem{Spelta_2020_NLMS}
M.~J.~M. Spelta and W.~A. Martins,
\newblock ``Normalized lms algorithm and data-selective strategies for adaptive graph signal estimation,''
\newblock {\em Signal Processing}, vol. 167, pp. 107326, 2020.

\bibitem{peng2023adaptive}
C.~Peng, Y.~Yan, and E.~KURUOGLU,
\newblock ``Adaptive message passing sign algorithm,''
\newblock in {\em Temporal Graph Learning Workshop @ NeurIPS 2023}, 2023.

\bibitem{Isufi2019gvarma}
E.~Isufi, A.~Loukas, N.~Perraudin, and G.~Leus,
\newblock ``{Forecasting Time Series With VARMA Recursions on Graphs},''
\newblock {\em IEEE Transactions on Signal Processing}, vol. 67, no. 18, pp. 4870--4885, 2019.

\bibitem{Hong_GGARCH_2023}
J.~Hong, Y.~Yan, E.~E. Kuruoglu, and W.~K. Chan,
\newblock ``Multivariate time series forecasting with {GARCH} models on graphs,''
\newblock {\em IEEE Transactions on Signal and Information Processing over Networks.}, vol. 9, pp. 557--568, 2023.

\bibitem{li2024simac}
Y.~Li, J.~Keung, Z.~Yang, X.~Ma, J.~Zhang, and S.~Liu,
\newblock ``Simac: simulating agile collaboration to generate acceptance criteria in user story elaboration,''
\newblock {\em Automated Software Engineering}, vol. 31, no. 2, pp. 55, 2024.

\bibitem{jang2023can}
J.~Jang, S.~Ye, and M.~Seo,
\newblock ``Can large language models truly understand prompts? a case study with negated prompts,''
\newblock in {\em Transfer learning for natural language processing workshop}. PMLR, 2023, pp. 52--62.

\bibitem{jia2024gpt4mts}
F.~Jia, K.~Wang, Y.~Zheng, D.~Cao, and Y.~Liu,
\newblock ``Gpt4mts: Prompt-based large language model for multimodal time-series forecasting,''
\newblock in {\em Proceedings of the AAAI Conference on Artificial Intelligence}, 2024, vol.~38, pp. 23343--23351.

\bibitem{Chen_2023_zero_shot}
J.~Chen, Y.~Geng, Z.~Chen, J.~Z. Pan, Y.~He, W.~Zhang, I.~Horrocks, and H.~Chen,
\newblock ``Zero-shot and few-shot learning with knowledge graphs: A comprehensive survey,''
\newblock {\em Proceedings of the IEEE}, vol. 111, no. 6, pp. 653--685, 2023.

\bibitem{maharana2024evaluating}
A.~Maharana, D.-H. Lee, S.~Tulyakov, M.~Bansal, F.~Barbieri, and Y.~Fang,
\newblock ``Evaluating very long-term conversational memory of llm agents,''
\newblock in {\em Proceedings of the Annual Meeting of the Association for Computational Linguistics}, 2024, vol.~1, pp. 13851--13870.

\bibitem{shahriar2024putting}
S.~Shahriar, B.~D. Lund, N.~R. Mannuru, M.~A. Arshad, K.~Hayawi, R.~V.~K. Bevara, A.~Mannuru, and L.~Batool,
\newblock ``Putting gpt-4o to the sword: A comprehensive evaluation of language, vision, speech, and multimodal proficiency,''
\newblock {\em Applied Sciences}, vol. 14, no. 17, pp. 7782, 2024.

\bibitem{karakucs2020modelling}
O.~Karakus, E.~E. Kuruoglu, and M.~A. Altinkaya,
\newblock ``Modelling impulsive noise in indoor powerline communication systems,''
\newblock {\em Signal, image and video processing}, vol. 14, no. 8, pp. 1655--1661, 2020.

\bibitem{lee2023deep}
W.~Lee, H.~S. Nam, J.~Y. Seok, W.-Y. Oh, J.~W. Kim, and H.~Yoo,
\newblock ``Deep learning-based image enhancement in optical coherence tomography by exploiting interference fringe,''
\newblock {\em Communications Biology}, vol. 6, no. 1, pp. 464, 2023.

\bibitem{herranz2004alpha}
D.~Herranz, E.~Kuruo{\u{g}}lu, and L.~Toffolatti,
\newblock ``An alpha-stable approach to the study of the p (d) distribution of unresolved point sources in cmb sky maps,''
\newblock {\em Astronomy \& Astrophysics}, vol. 424, no. 3, pp. 1081--1096, 2004.

\bibitem{kuruoglu1997new}
E.~Kuruoglu, C.~Molina, S.~Godsill, and W.~Fitzgerald,
\newblock ``A new analytic representation for the symmetric alpha-stable probability density function,''
\newblock in {\em Proceedings of the 5th World Meeting of the International Society for Bayesian Analysis (ISBA). ASA: American Statistical Association}, 1997, pp. 229--233.

\bibitem{kuruoglu2010using}
E.~E. Kuruoglu and A.~S. Taylor,
\newblock ``Using annotations for summarizing a document image and itemizing the summary based on similar annotations,'' May~4 2010,
\newblock US Patent 7,712,028.

\bibitem{saffarimiandoab2021insights}
F.~Saffarimiandoab, R.~Mattesini, W.~Fu, E.~E. Kuruoglu, and X.~Zhang,
\newblock ``Insights on features' contribution to desalination dynamics and capacity of capacitive deionization through machine learning study,''
\newblock {\em Desalination}, vol. 515, pp. 115197, 2021.

\end{thebibliography}
\end{document}